\documentclass[nohyperref]{article}

\usepackage{microtype}
\usepackage{graphicx}
\usepackage{subfigure}
\usepackage{booktabs} 

\usepackage{hyperref}

\usepackage[accepted]{icml2022}

\usepackage{amsmath}
\usepackage{amssymb}
\usepackage{mathtools}
\usepackage{amsthm}

\usepackage[capitalize,noabbrev]{cleveref}
\usepackage{xcolor}

\newcommand\eugene[1]{\textcolor{blue}{#1}}

\theoremstyle{plain}

\theoremstyle{definition}

\theoremstyle{remark}

\usepackage[textsize=tiny]{todonotes}
\usepackage[inline]{enumitem}

\icmltitlerunning{\hfill Machine Learning Systems Safety \hfill \thepage}

\begin{document}

\twocolumn[
\icmltitle{Concrete Safety for ML Problems: 

System Safety for ML Development and Assessment}




\begin{icmlauthorlist}
\icmlauthor{Edgar W. Jatho, III}{NPS}
\icmlauthor{Logan O. Mailloux}{NPS}
\icmlauthor{Eugene D. Williams}{NPS}
\icmlauthor{Patrick McClure}{NPS}
\icmlauthor{Joshua A. Kroll}{NPS}
\end{icmlauthorlist}

\icmlaffiliation{NPS}{Department of Computer Science, Naval Postgraduate School, Monterey, CA, USA}

\icmlcorrespondingauthor{Edgar W. Jatho III}{edgar.jatho@nps.edu}
\icmlcorrespondingauthor{Joshua A. Kroll}{jkroll@nps.edu}

\icmlkeywords{Machine Learning, Systems Safety Engineering, Artificial Intelligence, Safety, Social, Ethical, Responsible AI, Trustworthy AI, Fairness, Traceability}

\vskip 0.3in
]



\printAffiliationsAndNotice{} 

\begin{abstract}
Many stakeholders struggle to make reliances on ML-driven systems due to the risk of harm these systems may cause. 
Concerns of trustworthiness, unintended social harms, and unacceptable social and ethical violations undermine the promise of ML advancements. Moreover, such risks in complex ML-driven systems present a special challenge as they are often difficult to foresee, arising over periods of time, across populations, and at scale. These risks often arise not from poor ML development decisions or low performance directly but rather emerge through the interactions amongst ML development choices, the context of model use, environmental factors, and the effects of a model on its target.  Systems safety engineering is an established discipline with a proven track record of identifying and managing risks even in high-complexity sociotechnical systems. In this work, we apply a state-of-the-art systems safety approach to concrete applications of ML with notable social and ethical risks to demonstrate a systematic means for meeting the assurance requirements needed to argue for safe and trustworthy ML in sociotechnical systems.




\end{abstract}

\section{Introduction}

ML-driven systems promise new automation capabilities and stand to provide valuable new tools across a diversity of applications. At the same time, questions abound as to these systems' trustworthiness, safety, and traceability for failure cause. For this reason, ML system developers, practitioners and professionals who use them, along with industry and civil policymakers, and the citizens whose lives are affected by them, have all proffered interventions designed to build the ability to make reliances on ML-driven systems for particular purposes. However, to-date, the bulk of these interventions are either \emph{instrumental} changes at the \emph{component level}, which aim to improve particular performance metrics or specific aspects of ML development practice; or they are \emph{management and policy frameworks} (e.g., the new NIST AI Risk Management Framework) or \emph{lists of ethical principles}. One often-cited taxonomy of AI safety issues~\cite{amodei2016concrete} has been critiqued as taking an overly-instrumental frame~\cite{raji2020concrete}, a critique we extend to follow-on work focused on ML~\cite{hendrycks2021unsolved}. There is a substantial and long-recognized gap between improved component performance and systemic safety~\cite{perrow_normal_1984}; instrumental, componentwise improvement will not enable the safe use of ML. The latter two interventions, while taking a wider view of the problem, are often challenging to extract requirements from or bridge to technical decision-making. The bridging between high-level policy goals and ethical ideals and implementable technical decisions is critical to answering challenges to the utility of ML, particularly as new regulatory demands drive a need for assurance against failure in high-risk applications.\footnote{For example, the pending EU ``AI Act'' requires risk assessments of many ``high risk'' ML systems, and policy developments in the US could emerge quickly as the proposed ``AI Bill of Rights'' joins the newly released NIST AI Risk Management framework as a lens for social and ethical risks of ML-driven systems.}

We posit that the reason the ML research and practice communities struggle to assess safety and trustworthiness for ML-driven systems is that assessment is consistently performed only at the component level with technical assessments  centered on \emph{component-based properties}, such as: performance improvement and optimization; instrumental explainability of model input-output or feature structures; robustness improvements (including adversarial robustness); or simplistic fairness metrics. Given this component focus, it is no surprise that policy frameworks and associated risk tools struggle to assess risk at scale, over time, and across populations--these are \emph{system properties}. 
It is not that component level assessments are unhelpful, but that as yet, no systematic nor repeatable method to \textit{compose} such tools into assurance arguments has gained adoption. For sociotechincal systems, critical issues such as trustworthiness and social and ethical harms emerge at the system level; thus, it is at the system level that they must be reasoned about and addressed. This does not mean that ML researchers can build tools and leave these higher level issues as deployement concerns. Low-level data and ML decisions have significant impacts on whole-system properties. Decisions that optimize longstanding notions of component performance are often in tension with, or at direct odds with, minimizing social and ethical harms~\cite{martin_extending_2020}. 



Systems safety engineering provides well-studied tools, techniques, and procedures to identify and control risks in complex sociotechnical systems (collections of interacting organizational, human, and technical components)~\cite{shin_stpa-based_2021, leveson_engineering_2016}. Although others have suggested that safety frameworks can perform similarly for ML system risks, including social and ethical risks~\cite{dobbe_system_2022, raji_closing_2020, hendrycks_unsolved_2022}, 
the concrete use of these techniques remains unexplored, along with any investigation of whether they can surface social and ethical risks in particular. To test this hypothesis, we apply Leveson's System Theoretic Process Analysis (STPA)~\cite{leveson_stpa_handbook_2018} to two representative ML systems, giving special attention to socially and ethically problematic outcomes operationalized as safety hazards. 

Our first case study involves applying STPA to study prescription abuse risk scores in a notional Prescription Drug Monitoring Program (PDMP) modeling those used in a majority of U.S. states. Our second case study centers on an ML-based facial recognition system used in the criminal justice context. By analyzing the concrete application of STPA in realistic ML case studies, we can determine if STPA can effectively and repeatably identify social and ethical risks. Additionally, we are delivering reference architectures, which provide the analysis and structured reasoning to  proactively redesign elements to eliminate these risks and suggest specific means to monitor and mitigate them. 


\section{Background}

 The state of the art in ML evaluation generally relies on ad-hoc reviews of chosen metrics such as AUC, metrics derived from confusion matrices, or -- for social and ethical risks -- so-called ``fairness metrics''. Metric-based evaluation on its own is a narrow view of model performance, especially for social and ethical risks, as it frequently fails to address wider critical equities at stake~\cite{raji_ai_2021, malik_hierarchy_2020}. Fairness metrics, while a common proxy for identifying social and ethical concerns, are widely acknowledged to be imperfect operationalizations of underlying human values~\cite{mulligan_this_2019}. Additionally, it can be difficult to assign responsibility for social and ethical risks in ML systems or to determine appropriate interventions to mitigate problems, even once discovered~\cite{kroll_accountable_2017, raji_closing_2020, Cooper_2022, selbst_fairness_2019}.

In response to these shortfalls, several ambitious efforts aim to create practical, effective evaluation frameworks for identifying harms to mitigate social and ethical risks. For example, the U.S. NIST's AI Risk Management Framework~\cite{NIST_ai_2022} and the pending ``AI Act'' legislation in the European Union both categorize concerns about harm resulting from ML-driven systems as a risk management problem and envision solutions in standardized evaluation frameworks. However, these frameworks rely on  consensus best-effort and expert judgement; assessments of social and ethical risks remain ad-hoc even if systematized. Instead, effective risk governance must be based in scientific evaluation, thorough analysis, and process validation. Practitioners and academics alike recognize the need for robust evaluation practices and would welcome such a substantiated standardized framework~\cite{holstein_improving_2019, madaio_assessing_2022, rismani2023airplane, wilson_building_2021}.


\subsection{Component Reliability is \textit{Not} Safety}
Historically, various methods for identifying and mitigating risks in complex systems sought to do so using two general approaches: 1) reducing the probability of component failures, and/or 2) reducing the severity of the effects of each particular component's failure.  The logic driving these approaches being that, first, lower failure rates of constituent components should necessarily lead to lower system failure rates and, second, lower severity of component failure, often through the introduction of redundancy, should likewise lead to less severe overall system failures.  Perrow, however, posits that some systems are so complex and/or tightly coupled that even in the presence of (and not uncommonly \emph{because of}) redundancies and fail-safes, they can and do experience unforeseen catastrophic interactions between simultaneous component failures that, though estimated to be of vanishingly small probability, in fact commonly co-occur. He holds that far from being unlikely, accidents of this kind are inevitable, titling his book, ``Normal Accidents''~\cite{perrow_normal_1984}. Leveson extends this concept and points out that component failures or human errors are \textit{not required} for a failure or accident to occur, but instead system failure may result from unforeseen interactions between \textit{properly functioning components}~\cite{leveson_moving_2009}. Such interactions and subsequent failures are \textit{emergent} system behaviors, arising from inadequate or unsafe control actions in a system's control structure~\cite{leveson_engineering_2016}. Examples of this phenomenon abound: consider the Arianne 501 explosion~\cite{lions1996ariane} or the Mars Polar Lander crash~\cite{albee_polar_report_JPL}, both driven not by compounded component failures but by control failures far from the proximate causes of loss. Methods that only consider components and their failures \textit{cannot} foresee system failures of this kind.

A strength of system safety engineering frameworks as a whole is that they connect abstract safety policies to implementable requirements, which are often difficult to make actionable through technical means alone. Tools from this discipline further have the advantage of being regularly applied in consequential domains, being well studied, and providing a strong basis on which to systematize efforts to identify social and ethical risks~\cite{antoine_systems_2013, rising_systems-theoretic_2018}. Component-level techniques include traditional safety-through-reliability techniques like fault-tree analysis (FTA)~\cite{lee_fault_1985} and Failure Mode and Effects Analysis (FMEA), both quantification-oriented approaches used for decades to reduce the number and frequency of failures in systems under analysis~\cite{stamatis2003failure}. 

\subsection{Systems-Theoretic Process Analysis (STPA)}
Leveson's Systems-Theoretic Accident Model and Process (STAMP) body of tools~\cite{leveson_engineering_2016} explicitly rejects the notion that reducing failures necessarily improves safety, noting that safety is a property of systems, not components. STAMP models sociotechnical systems in terms of control and control structures. In this paradigm, unwanted losses and hazardous system states are understood as the result of insufficient control within the system.

As part of STAMP, Systems-Theoretic Process Analysis (STPA) is a top-down systems safety analysis approach with a successful track record in high consequence domains, from nuclear power plants to space flight~\cite{shin_stpa-based_2021, ishimatsu_hazard_2014,leveson_engineering_2016}. STPA is defined by four sequential, yet recursive steps: 
\begin{enumerate*}
\item Defining the purpose of the analysis, including defining unacceptable stakeholder losses and hazardous system states;
\item Modeling the system's control structure including control actions and feedback loops;
\item Identifying unsafe control actions (UCAs) within each control loop which can result in a hazardous system state or unacceptable loss; and
\item Identifing and considering loss scenarios that lead to UCAs to ensure hazardous system states are properly addressed.
\end{enumerate*}

By considering the full sociotechnical system, STPA can contextualize ML hazards with respect to identified social and ethical losses even when those undesired behaviors arise from interactions between technical components, humans, organizations and environmental factors. Recognizing and responding to the social and ethical risks of an ML model requires viewing that portion of the system in its context of use, as part of its broader sociotechnical system and in light of its control structure~\cite{martin_extending_2020}. 


\section{Methodology}\label{methodology}
\begin{figure*}[h!!]
\centering
\includegraphics[width=0.99\linewidth, height=3.5cm]{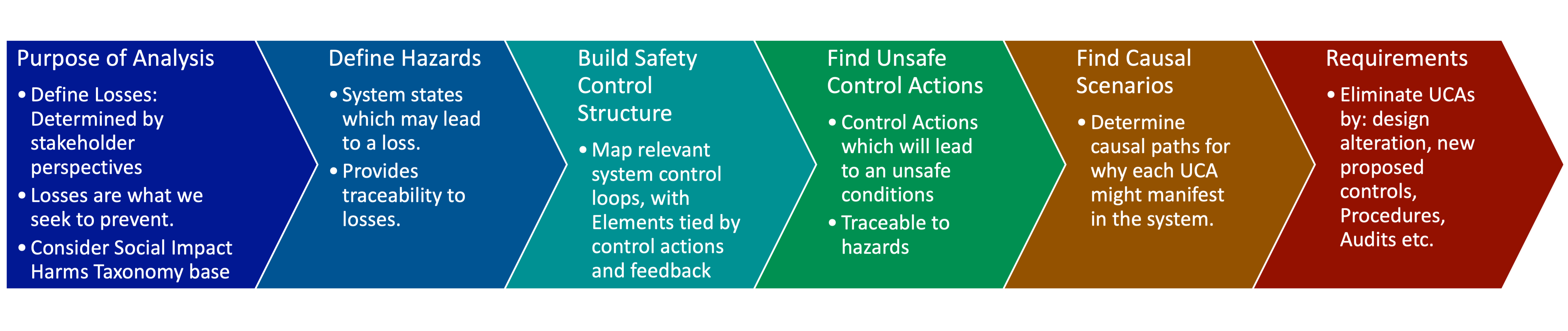} 
\caption{STPA six step overview}
\label{STPA}
\end{figure*}

Figure 1 presents a six step overview of the STPA process. Walking through the STPA method, we begin by defining the purpose of our analysis: defining stakeholders, understanding the intended system and its existing sociotechnical context, setting the scope of analysis, and understanding the kind of losses we want to prevent. It is also important to understand at this stage that losses can, and frequently do, exist outside the scope of the system of interest (SOI). However, STPA is designed to address losses wherever they occur, and modeling when losses occur beyond the system boundary can help define new system requirements to control the effects of the loss or may suggest broadening the scope of analysis in order to effectively control the loss. 



We next capture and define the system's hazardous states. Hazards are system states which could lead to a defined loss. In each analysis, we created a substantial initial hazard list, and revised it during each later step of STPA by consolidating and refining hazards. It is relatively easy to generate rather large lists of potentially hazardous systems states, so the analysis team must constantly be reminded to only go as low (in abstraction) as necessary to reason about controlling said hazards to prevent unacceptable losses from occurring. High-level hazards may be decomposed into detailed hazards as the model is enriched to more detailed levels of abstraction.

Next, we build out high level mappings of the system control structure with linkages to important contextual elements, see figures \ref{FRT_ctrl} and \ref{PDMP_ctrl} for detailed examples. In this step, control loops are modeled where the controller and its controlled process are tied together by \textit{control actions} from the controller and \textit{feedback} from the controlled process. For the purposes of our investigation, we scoped our analysis to focus on two or three of the most interesting ML related control loops as discussed in detail below.

The STPA analysis proceeds by analyzing \textit{unsafe control actions} in each control loop. These unsafe control actions (UCAs) are found by identifying necessary control actions between the controller and the controlled process and then applying a slightly modified (as described in Section~\ref{sec:se-stpa}) STPA rubric to uncover paths to hazardous system states. That is, for each control action we ask whether any of the following cause the system to enter a hazardous state: not providing the control action; providing the control action; executing the control action too early, too late or in the wrong order; or providing the control action to excess or insufficient degree. Using this rubric, we build a UCA analysis table which identifies all of the hazardous system states for each misapplied control action.

Lastly, we examine loss scenarios, which are causal paths leading the system into hazardous states. A number of loss scenarios are proposed and then walked through the control structure to examine why a UCA might occur or why an otherwise safe control action might lead to a hazard. Detailed analysis at this step also ensures measurable criteria are attached to each control action and feedback loop. Additionally, this step serves as a verification and validation check of the UCAs to ensure the analysis is complete and each UCA is correctly mitigated. See \ref{CJFR_UCA1} for an example of a completed UCA analysis table. 

Once finalized, the UCA analysis table contains a validated listing of UCAs providing traceability to hazardous system states and stakeholder-defined unacceptable losses that can result. The STPA process not only identifies system hazards, but describes them in a way that leads directly to offsetting control requirements to avoid losses. Moreover, these requirements are solution neutral: they express the need for better control but not how to effect that control. UCAs can be addressed in the most effective and efficient way suited to the system. For example, in some cases it may be desirable to redesign the system to eliminate the risk entirely, while in other cases it may be desirable to institute new controls to monitor the risk, and in other situations the only solution maybe to provide supplemental training to avoid the risks.

\subsection{Team Composition and Study Limitations} \label{team}
Our initial PDMP STPA analysis team was composed of five ML researchers, one sociologist, and one STPA subject matter expert. For the second case study, the criminal justice facial recognition system, our team was composed of four ML researchers and one STPA expert. Four members of the team participated in both analyses, including the primary researcher, our responsible ML researcher, and our STPA expert. Notably, the involvement of our sociologist researcher paid dividends in both STPA analyses. Both teams were adequate for the analysis performed.\footnote{Because of the uniquely conceptual nature of STPA, it is worth stating that the STPA analysis and results are inherently limited by the Subject Matter Experts (SMEs) included on the STPA team, much more so than a conventional safety analysis effort.} 
In addition to working with an experienced STPA expert, each member of the team worked to become thoroughly familiar with STPA by reading core STPA expositions -- Leveson's \textit{Engineering a Safer World} textbook~\cite{leveson_engineering_2016}, studying the \textit{STPA Handbook}~\cite{leveson_stpa_handbook_2018}, and watching training videos~\cite{thomas_2021}.

We did not have direct access to a PDMP scoring system or its datasets. However, we were able to derive necessary details from publicly available operating manuals, pharmacy work-flows, state and legal references, and training documents. Additionally, we relied on recent research, which re-creates and tests ML models using similar datasets and features, as are known to be used in the real systems to be very useful. State manuals describe key variables as ``predictive of unintentional overdose death'' was very insightful to understand optimization targets and extracted features~\cite{Kilby_algorithm, indiana_narx_manual, texas_narx_manual,NC_narx}.

Likewise, we did not have direct access to a law enforcement version of the subject facial recognition system or its datasets.  However, we were, again, able to derive necessary details from patents, company descriptions, legal requirements, and government testimonies. Additionally, we were able to study open source API documentation and request limited access to the system for evaluation purposes.

Given the top-down framing of STPA, these artifacts proved sufficient to provide ample sociotechnical context, operational details, and developmental insights to perform an effective exploration of STPA for social and ethical impacts on the two case study systems.

\section{Case Studies}
\subsection{PDMP Abuse Risk Scoring System}\label{PDMP}

Prescription Drug Monitoring Programs (PDMPs) are mandated in all 50 states and are intended to prevent or curtail widespread healthcare issues such as drug addiction, misuse, and overdose deaths~\cite{narxcare_webpage}. A majority of these programs require physicians, pharmacists and their staffs to score patients for drug misuse risk during clinical interactions such as writing or filling prescriptions for certain schedules of drugs~\cite{west_virginia_2018, indiana_narx_manual}. Risk scores are calculated by ML systems trained on a variety of data sources~\cite{texas_narx_manual,kilby_economics_2016}. Thus, these ML-based PDMP systems assist in governing the health care of hundreds of millions of people across the United States.\footnote{The authors performed an informal search of state public health websites and official news releases and confirmed 27 of 51 states (including the District of Colombia) use an ML-based scoring system as a major component of their PDMP. Additionally, five of the top seven U.S. pharmacy businesses (by 2021 prescription revenue)~\cite{fein_pharmacy_2021} require their pharmacists to use an ML-based PDMP scoring system in their workflow. Many of these entities use the same scoring tool provided by a single third-party vendor.}

\subsubsection{PDMP ML STPA Analysis}
Our PDMP control structure is shown in Figure 2 with a subset of key stakeholders. In systems like PDMP, which affect large numbers of people in highly consequential ways, impacting life, health and livelihood, it is incumbent on developers, company management, and government officials to demonstrate due diligence by showing evidences that the ML-enabled system not only improves the performance of the sociotechnical system which it is augmenting, but that it does not also introduce unacceptable negative social and ethical impacts~\cite{kroll_accountability_2020, raji_ai_2021, martin_participatory_2020}. The seriousness of this responsibility becomes even more critical when a single vendor provides the majority of services to an entire country. Further, when that single vendor augments both public and private decision making, the score's systemic impact rises, warranting additional scrutiny. However, social impact analysis is often done only instrumentally (e.g., examining the dependence of the score on protected attributes) or extensionally (e.g., examining outcomes only at the whole-system level). Such analysis is often ad-hoc, or is treated as if it were impossible due to its complexity~\cite{martin_extending_2020}. However, once the STPA process has been executed fully(see appendix \ref{execute_pdmp_stpa}), and UCAs are captured in analysis tables (e.g., table \ref{UCA_dev}) and linked to hazards, developers have a ready made list of actionable sociotechnical system requirements for fair and ethical treatment complete with traceability back to stakeholder-defined unacceptable losses. Finally, examining loss scenarios, and, thus, each UCA, helps system owners, operators, and developers systematically eliminate or mitigate hazardous systems states tied to identified harms.

\begin{figure}[h!]
\includegraphics[width=0.9\linewidth, height=6cm]{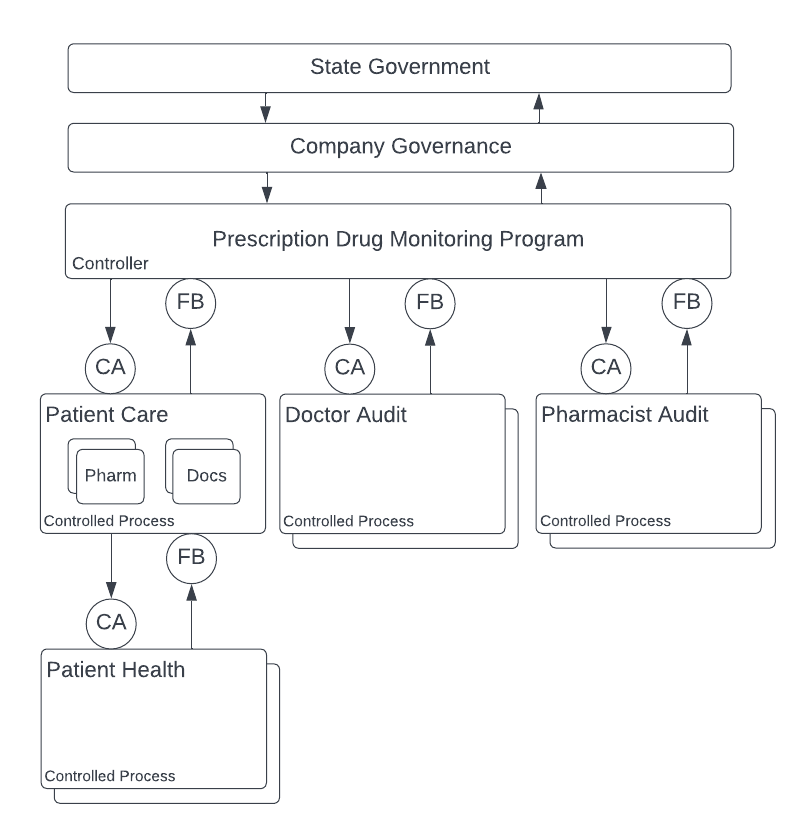} 
\caption{Prescription Drug Monitoring Programs (PDMP) High Level Control Structure}
\label{PDMP_ctrl}
\end{figure}
\subsubsection{PDMP Findings}

Many calls for action to adopt PDMP risk scores were motivated by the U.S. opioid epidemic -- the tragic and increasing loss of life fueled by opioid overdoses. One important fact that our analysis highlighted was the gap between stated goals for the system (reducing Opioid Use Disorder and preventing overdoses) and the optimization target for the score (minimizing drug \textit{diversion}, the illegal distribution or abuse of prescription drugs or their use for purposes not intended by the prescriber). The use of this proxy measure depends on the assumption that reducing drug diversion saves lives; however, when considering the hazards, control loops, and UCAs we identified, we conclude that the opposite relationship could be an unintended outcome of the system. Indeed, nationwide statistics show: 1) a marked reduction in drug diversion and opioid prescribing coinciding with the adoption of PDMPs in every state; and 2) an increasing overdose death rate overall~\cite{NIH_overdose_2022}. A frequent problem in ML systems is a mismatch between the objective function and the system goal~\cite{obermeyer_dissecting_2019}. Although the mismatch was previously identified in this system, STPA immediately raised this issue as a possible hazard in the first steps of the analysis and further developed the consequences this hazardous system state as the analysis proceded. By contrast, prior identification of the problem required substantial research efforts.

Second, the STPA control loops reveal mechanisms whereby actions taken by prescribers and dispensers which ostensibly protect an individual have the potential to drive patients to more desperate and dangerous behavior while weeking safe and effective treatment not provided through ordinary channels. For example, the aggregate total of providers' patient load scoring creates pressure on a physician or pharmacist which affect individual-level care decisions unrelated to the aggregate. Likewise, the system may bias providers towards extreme risk aversion, denying what maybe the most effective treatment to patients to protect their own ratings.

Finally, our STPA highlights the losses introduced by the PDMP risk score resulting from high false positive and high false false negative rates. Kilby found that systems based on the same datasets and features used in the most ubiquitous PDMP risk score systems, were shown to have prohibitively high false positive and false negative rates~\cite{Kilby_algorithm}. Hazards from high error rates remain even when considering the patient population with the highest 1\% of risk scores. Of this group, only 11\% are true positives. In the same top-1\%-by-risk-score population, the model only identifies 37\% of the total true positive patients who require the intervention the score is meant to help target~\cite{Kilby_algorithm, kilby_economics_2016}. For any ML system whose effects are felt at scale, two take-aways immediately follow from this analysis. One, it is relevant and important to consider how often identified hazardous states are entered, and two, structure the surrounding sociotechnical system to compensate appropriately in order to mitigate foreseeable losses. STPA provides a systematic means for identifying and capturing these seemingly obscure facts and coupling them with measurable and auditable mitigation and elimination strategies. In this way, STPA can serve as a means for providing a standard for the minimum necessary due diligence to field ML-driven high consequence sociotechnical such systems.

\subsection{Criminal Justice Facial Recognition (CJFR) System}
As of 2016, over 3,200 state and local law enforcement agencies reported leveraging facial recognition technology to assist in criminal investigations, and by 2020 at least 20 federal agencies reported using face recognition for law enforcement purposes~\cite{garvie_forensic}. In testimony before the U.S. House Oversight and Reform Committee, it was reported that the FBI Criminal Justice Information Services division received 152,565 facial recognition search requests to its NGI-IPS repository\footnote{Next Generation Identification - Interstate Photo System.} between December 2017 and April 2019~\cite{testimony_FBI_facial}. This total averages to almost 9,000 facial recognition identification requests per month. However, at this time there is no U.S. standard guidance for the integration and use of AI or ML-based facial recognition technology in law enforcement settings~\cite{NIST_facial_2023}. \footnote{The FBI reports having its own standard procedures for using facial recognition in investigations. Training includes manual reviews of matching probe and candidate photos.~\cite{testimony_FBI_facial}} 

Currently, there is no overarching governance process or standardized guidance for the immensely consequential use of ML-driven facial recognition algorithms in complex sociotechnical law enforcement structures. This is troubling given the number of agencies using these types of technologies as inputs to life-impacting, high-consequence decisions. As a result facial recognition policies, procedures, and implementations vary within law enforcement agencies. In the best case, a trained facial recognition team regularly probes the database and verifies results; in other deployments, lone untrained local law enforcement investigators use a facial recognition systems with little or no training~\cite{garvie_forensic}. 

Lastly, it is worth noting that one industry-leading U.S. company provides contracted support to the FBI, U.S. Immigration and Customs Enforcement, and the Department of Defense. Moreover, they boast over 30 billion photos in its facial recognition database--representing the equivalent of nearly four photos for every human on earth~\cite{brodkin_clearview_2022, USAspending_clearview}. These photos are harvested from the internet, social media, news articles, company websites, and other available data sources. It is also worth noting that, as of 2016, it was estimated that half of all American adults were already in law enforcement facial recognition systems. In querying these photos, the above company also reports an astounding ``99\%+ accuracy across all demographics" when seeking to match photos to identities according to NIST testing standards~\cite{NIST_FRVT_1_verification}. An example of this ML-driven facial recognition system is shown in Figure 3.
\begin{figure}[h!]
\includegraphics[width=0.9\linewidth, height=6cm]{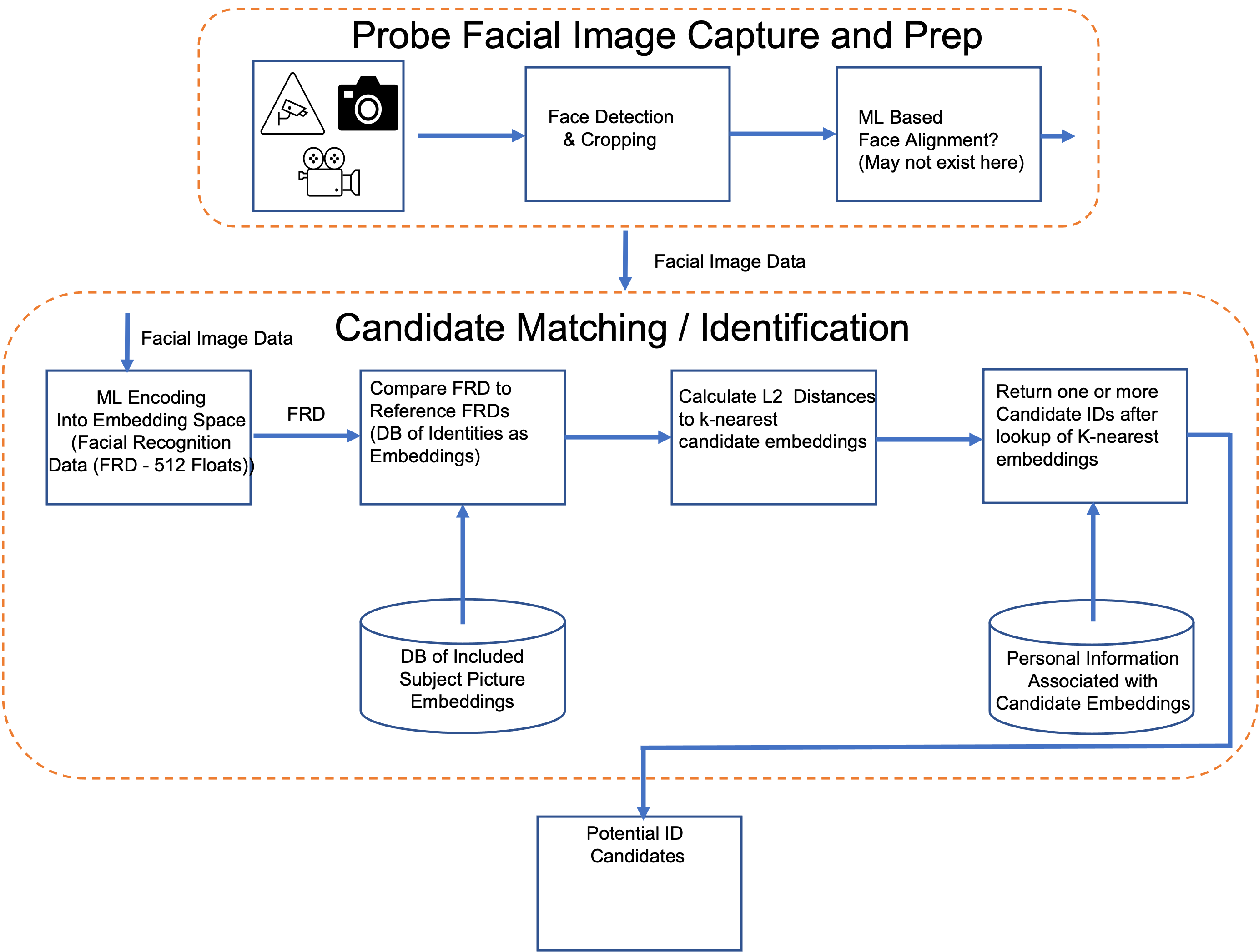} 
\caption{Notional Facial Recognition System}
\label{FRT_alg}
\end{figure}

\subsubsection{CJFR ML STPA Analysis}

Because of the complexity associated with performing social and ethical STPA for a ML-driven sociotechnical CJFR system, we found it helpful to categorize our losses into two broad types: individually experienced losses and losses that can only be described at the societal level. We found this categorization useful to carry through the STPA process, separating hazard analysis and the consideration of loss scenarios as they dealt with different parts of the larger CJFR system. Moreover, it was advantageous to work with smaller numbers of focused losses rather than a large, disparate list. A full list of defined losses can be found in Appendix \ref{frt_losses}.

Next, we identify 21 hazardous system states, mapping each state to one or more of the related losses, see table \ref{frt_hazards} in the appendix. In figure \ref{FRT_ctrl}, we depict a control structure for the notional criminal justice facial recognition system. Analysis of the CJFR control loops resulted in a combined 32 unsafe control actions (UCAs) as shown in tables \ref{CJFR_UCA1}, \ref{CJFR_UCA2}, and \ref{CJFR_UCA3}, respectively. Once the UCAs are captured in the analysis table and linked to possible hazards, the development team has a tailor-made scaffolding to construct  sociotechnical safety requirements and to reason about and consider implementation solutions in a structured way.

For example, consider a loss scenario where a facial recognition ML was used to establish probable cause, leading to a warrant for the arrest of someone who was not the true perpetrator of a crime but merely someone who looked like a photograph which was wrongfully corroborated on by an unaware witness. Our analysis yields this sort of scenario mechanically, without the need for unstructured reflection and simultaneously suggests opportunities to intervene on the system and avoid the loss.

\begin{figure}[h!]
\includegraphics[width=0.9\linewidth, height=6cm]{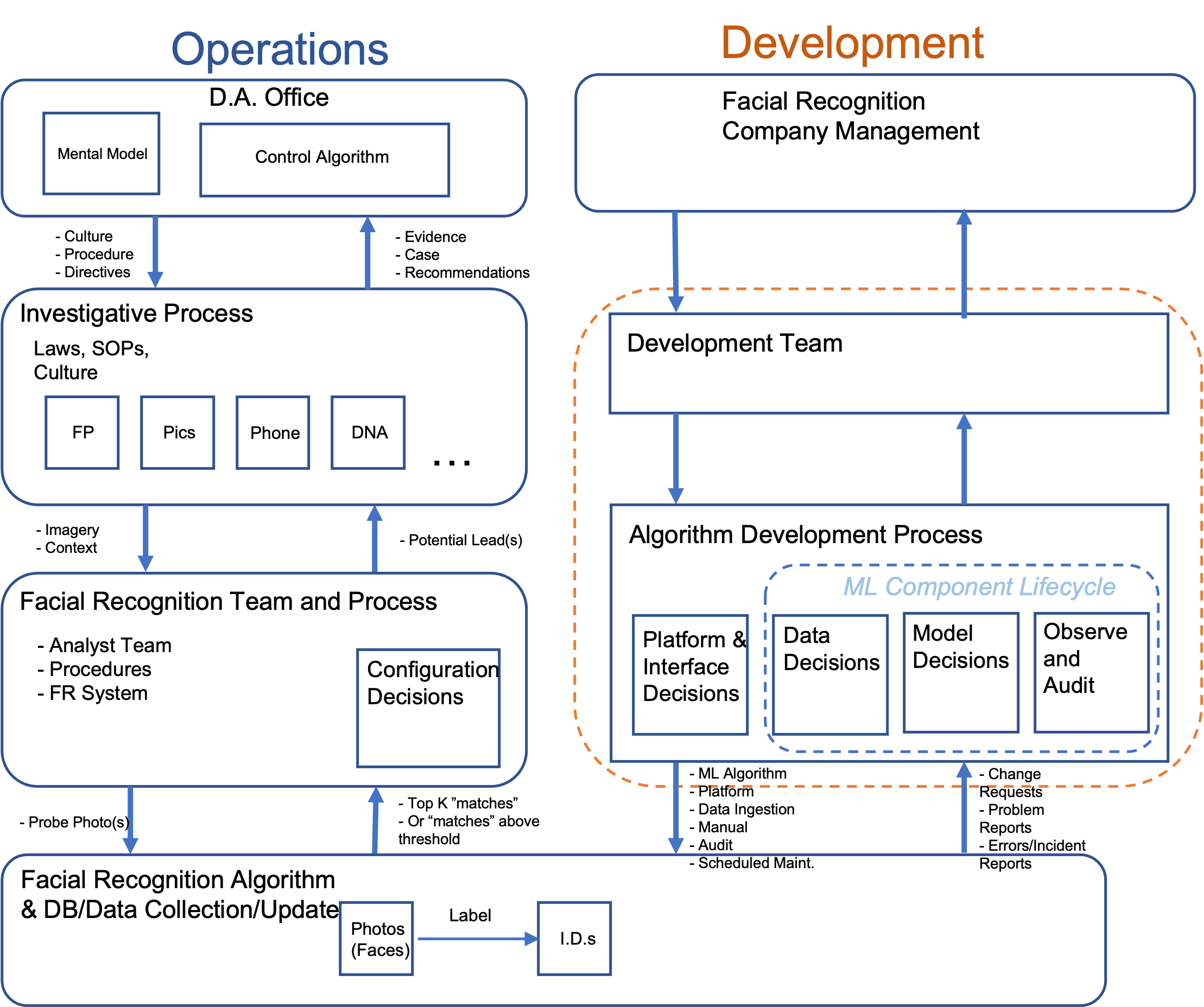} 
\caption{Notional CJFR Control}
\label{FRT_ctrl}
\end{figure}

\subsubsection{CJFR Findings}
In developing the CJFR control structure, we quickly settled on a bilateral design separating an operational control structure from a development control structure, each ``controlling'' the facial recognition algorithm at bottom. On the development side of the structure, STPA structures reasoning about machine learning life-cycle decisions. We model the development team as a controller and three life-cycle phases as controlled processes: \textit{Problem Framing and Data Decisions} (PFDD); \textit{Modeling Decisions}; and \textit{Deployment, Observation, and Audit}. In this framing, \textit{decisions} made by the development team represent control actions.

Our analysis of the CJFR system focuses on the development control loops; our goal is to structure reasoning about problems that arise from those decisions and to monitor and adjust development decisions to mitigate (or in the best case, eliminate) social and ethical concerns operationalized as system hazards. One benefit of framing the process in this way is that it casts every decision in the development cycle in terms of its effects on social and ethical impacts downstream. For instance, one control action in the PFDD control loop is to ``define social and ethical success metrics." \textit{Not providing} success metrics that consider social and ethical impacts is then an unsafe control action which results in hazards even at a high abstraction level (e.g., wrongful warrant issuance or improper investigator confidence in results) and at the lower abstraction levels (e.g., failing to consider social and ethical impacts when evaluating the balance between false matches and false non-matches or examining even small performance differences between demographics).


Finally, When applying facial recognition systems across large populations, such as the U.S. or even say a city such as New York, even a 99.9\% accuracy across the population may not be enough. If such a system was, for example, used on each of approximately 5.4 million adults in New York City, 5,400 people, on average, would be misidentified. Likewise, claiming all demographics are above 99\% accurate and thus sufficiently equal, also may not be enough to prevent significantly harmful disparities. Although these failure modes are well known, we stress that STPA offers a repeatable process to identify, discuss, and intervene on problems that arise at the whole-system level, and enables prospective analysis at the design stage expanding on componentwise performance evaluation to create holistic claims of assurance.

\section{Social and Ethical Impact Analysis for ML}
\label{sec:se-stpa}
Fundamentally, the ideas driving STPA are readily understandable, however, performing conceptual analysis can be challenging for ML-driven systems as these systems do not have traditional software specifications but rather induced behavior from decisions about data and modeling choices. This is especially true with regard to conducting social and ethical impact analysis where the manifestation of the losses are often spread across a wide variety of stakeholders and difficult to localize to component misbehavior. Thus, the analysis team can tend to spend considerable time working through competing levels of abstraction and multiple variations of STPA losses, hazards, control actions, and feedback loops. Moreover, if the problem is not carefully circumscribed, the team can struggle to understand the right level of ``controller'' and ``controlled process'' problems to study. Since the benefit of modeling is often realized in working out unspecified details, these exercises enable the creation of ersatz specifications from which assurance arguments can be derived and paths to defined failures elucidated.



\subsection{Social and Ethical Analysis Considerations}
In our case studies, a number of the controllers or controlled processes involve humans with substantial discretion, such as doctors, pharmacists, and their staffs.  Goodhart's law, ``when a measure becomes a target, it ceases to be a good measure'', implies for our analysis that the behavior and effectivenss of controls may change over time. One way that structured analysis such as STPA aids social and ethical risk assessment is by highlighting where this can happen and concretely identifying its consequences. For example, in our PDMP score analysis, such feedback-driven system drift describes undesired behaviors: abandonment by doctors and service refusal by pharmacists, both of which lead to attendant losses. Identification of these behaviors expands the potential to identify hazards and loss scenarios, leading to the identification of new controls to prevent those hazards.

\subsection{ML Analysis Benefits}
Conducting an STPA for social and ethical impact forces development teams and other stakeholders to do the necessary work to consider the larger sociotechnical system a model inhabits. Frequently, ML teams are hyper-focused on performance measures and thus miss compositional effects~\cite{raji_ai_2021}. The STPA approach cultivates a rich appreciation for the complexity ML-driven solutions are joining 
and provides an effective model with which to reason about harms, as well as the instruments, tools, and methods (both procedural and technical) that we can bring to bear to eliminate or mitigate them.

Another value in using STPA is the mandate to consider carefully the overall goal of the system. In our treatment, this led us to verify that the objective function adopted does not create a mismatch between model behaviors and system goals. For example, applying STPA to the health benefits scoring system studied by Obermeyer~\cite{obermeyer_dissecting_2019} could reveal the need for a check for racial and socio-economic disparities resulting from an objective function mismatch early in development and could have enabled a shift to a more appropriate governing optimization. A similar truth may hold for PDMP, more attention should be given to whether the current optimization drives the system in the direction of the overall ostensible goal.  

A specific contribution of this paper is a useful demonstration of the value of an abstraction of the machine learning lifecycle for use with STPA which recovered potential unsafe control actions during model development in a manner similar to those captured by their analogous operational control loops. Finding a way to abstract, model, and study the ML lifecycle for systemic impacts was a challenging aspect of applying STPA. However, the benefits of the approach are many. One benefit of the ML life-cycle treatment the team developed is that it breaks the STPA modeling into manageable pieces where each phase or stage has a reasonable number of control actions which can be more easily understood across various disciplines and used for multiple purposes from responsible development to auditing. Second, this approach allows the model components, control actions, and feedback loops to be further decomposed into measurable and verifiable considerations and checks. Lastly, this work can serve as a reference model for the community to study and expand.

One final benefit of the STPA process is that it necessarily provides a traceable path from every derived requirement to its contributing unsafe control actions, feedback loops, and orginating stakeholder defined losses. This property enables the complete retracing of the reasoning behind every decision in design, development, and operation.

\subsection{Considerations For Systems at Scale}
An important characteristic of ML systems is that they are applied at scale and encounter populations distinct from those for which they are designed and evaluated. It is necessary to control for this generalization and drift. Practical mechanisms for such control often rely on ongoing measurement of system performance or aggressive retraining, ad hoc solutions to a problem where issues are considered too difficult to foresee. The structured approach of STPA enables the sort of prospective analysis at the control level to define the risks and benefits of various control approaches and avoid the need for unbounded operational decision envelopes.

\section{Conclusion}
In this paper, we record the results of two case studies in which we applied STPA, a traditional system safety engineering analysis methodology to the challenge of assessing the social and ethical impact of two very different high-consequence ML-driven systems: 1. a Prescription Drug Monitoring Program \textit{risk score} in the healthcare context and 2. an ML-driven facial recognition system in the criminal justice context. We found that STPA's rigorous approach resulted in a thorough understanding of each subject systems' sociotechnical control structure and context as well as the necessary language and reasoning structure with which to build an assurance case that an overall sociotechnical system exhibits the requisite presence of an emergent property, such as trustworthiness or safety.  The process reveals a trove of hazards and unsafe control actions against which new system requirements for sociotechnical control mechanisms could subsequently be applied to prevent social and ethical losses.

We believe these case studies demonstrate that STPA provides a promising path to recover potential social and ethical losses, hazardous states, and unsafe control actions--and thus provide developers with a proven methodical, repeatable and effective approach to analyzing potential and existing ML-driven sociotechnical systems for social and ethical impact.
Future work will further examine the applicability and generalizability of STPA for ML-driven social and ethical impact by investigating a case study involving a large language model (LLM) driven content moderation system for social media.


\nocite{langley00}

\bibliography{main}
\bibliographystyle{icml2022}

\newpage
\appendix
\onecolumn
\section{Notes}
\begin{itemize}
    \item One paragraph distillation of Normal Accidents - Perrow --If they are so ``normal", then they are identifiable and solvable.
    \item Why the appropriate abstraction level to analyze and address these issues is the system level, versus the component level.
    \begin{itemize}
      \item (behavior analysis) an aspect of STPA is that of modeling and understanding behaviors - how a controller influences a particular outcome by applying a set of actions. 
      \item (component interactions) since one aim is to allay sociotechnical concerns (those that pertain to how systems and humans work together), its necessary to examine how the interconnected components of the system might lead to harms which, otherwise, are not immediately discernible in a piecewise analysis
      \item (proper failure detection) unlike traditional hazard analysis where we seek to identify how the failure of a singular component resulted in the failure of the system, we understand that ethical hazards and losses are not, generally, rooted in the control actions of a single component, but in how feedbacks are interpreted and acted upon.
    \end{itemize}
    \item Trustworthiness, safety, and ethical behavior are system-level properties of complex sociotechnical systems.
    \item A metric does not constitute an assurance argument. 
    \item There has been alot of work on fairness, explainability, performance metrics and causal methods, but there has not been sufficient attention paid in ML circles as to how and what methods can be employed to repeatably and reliably compose those methods and metrics such that they respond to concerns about safety, trustworthiness and ethics.
    \item Motivate STPA as the standard for system level safety analysis. 
    \item Accepting a complex systems viewpoint still leaves you with problem of how to take engineering practices and capture
    
\end{itemize}

\eugene{
        \begin{itemize}
            \item (Scenario) A warrant is issued for a person who is provided as a candidate match by the facial recognition system. They closely match the appearance of a suspect, but they are not the actual perpetrator.
            \item (UCA - control action leading to unsafe condition) Problem Framing and Data Decision ML model development phase where the metrics defining ``success" for the system do not take into account social and ethical impacts (what ``Good" means when measuring performance of the model's facial recognition function)
            \item (Hazard) Investigator maintains a false (improper) sense of confidence in system results - This can be caused by the development team not providing success metrics that consider social and ethical impacts - the social impact of falsely accusing someone innocent of an crime; This is worsened when the metrics defining accuracy (measurements or metadata accompanying results) are not providing the needed transparency to inform the system user (Investigation team) of the potential flaws or deficiencies in the system's functions which would, otherwise, lead a user to question the trustworthiness of the predictions/recommendations
            \item (Losses - not exhaustive) Societal: Loss of Justice, Loss of Opportunity, Individual: Loss of Confidence in Justice system
        \end{itemize}
    }

\section{Control Structure}\label{figs}
Figure \ref{PDMP_ctrl} situates our identified operational control loops for the PDMP scoring algorithm within the health system. Figure \ref{care} shows more detail for the particular Patient Care control loop for the PDMP score. Figures \ref{cycle}, \ref{cycle-treatment} and \ref{data} show the machine learning lifecycle, how this study proposes to model that cycle as a set of control loops and the specific control loop addressing the Problem Conception and Data Decisions portions of the cycle, respectively.  
\begin{figure}[h!]

\includegraphics[width=0.49\linewidth, height=6cm]{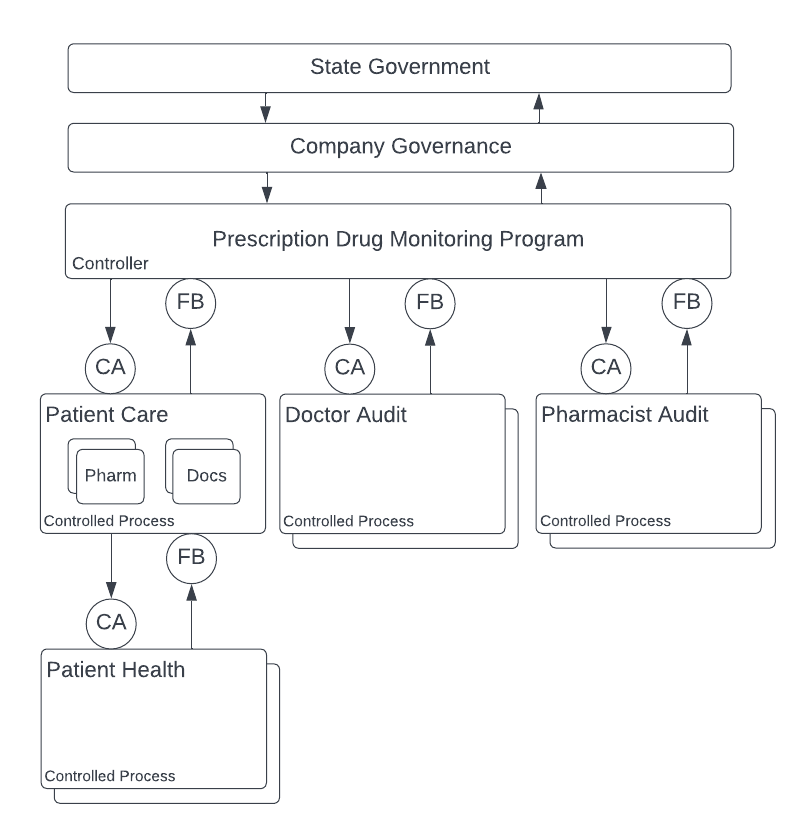} 
\caption{High Level Control Structure}
\label{PDMP_ctrl}
\end{figure}

\begin{figure}[h!]
\includegraphics[width=0.49\linewidth, height=6cm]{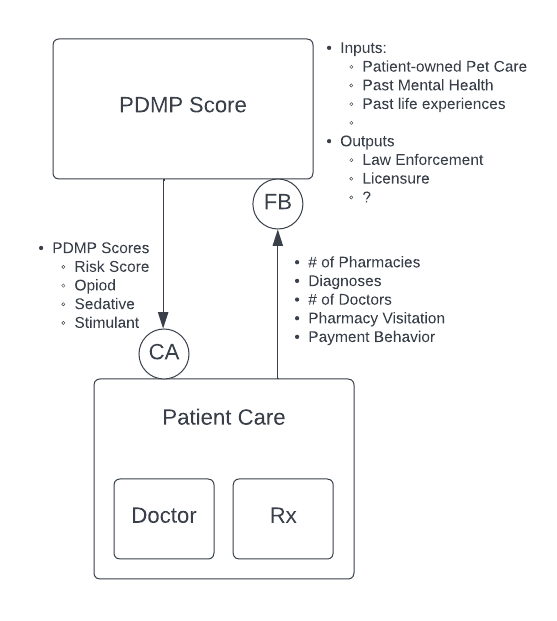}
\caption{Patient Care Control Loop}
\label{care}
\end{figure}

\begin{figure}[h!]
         \centering
         \includegraphics[width=0.49\linewidth, height=6cm]{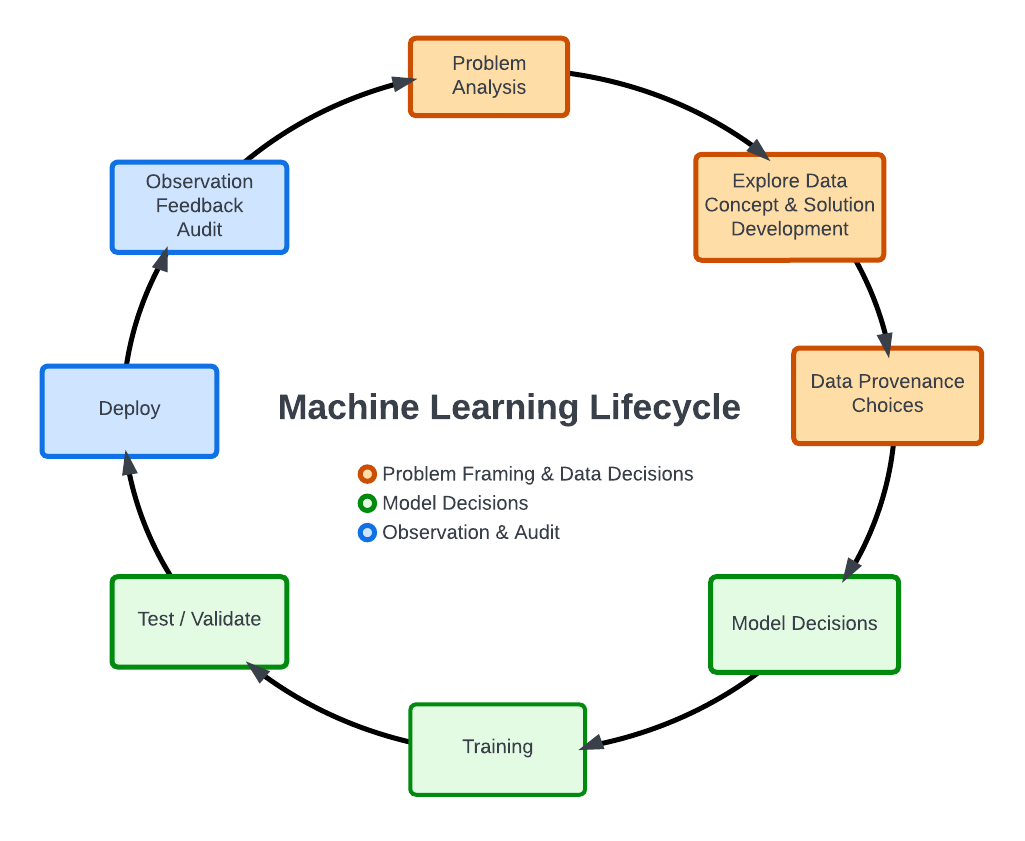}
         \caption{ML Life-Cycle}
         \label{cycle}
\end{figure}

\begin{figure}[h!]
         \centering
         \includegraphics[width=0.49\linewidth, height=6cm]{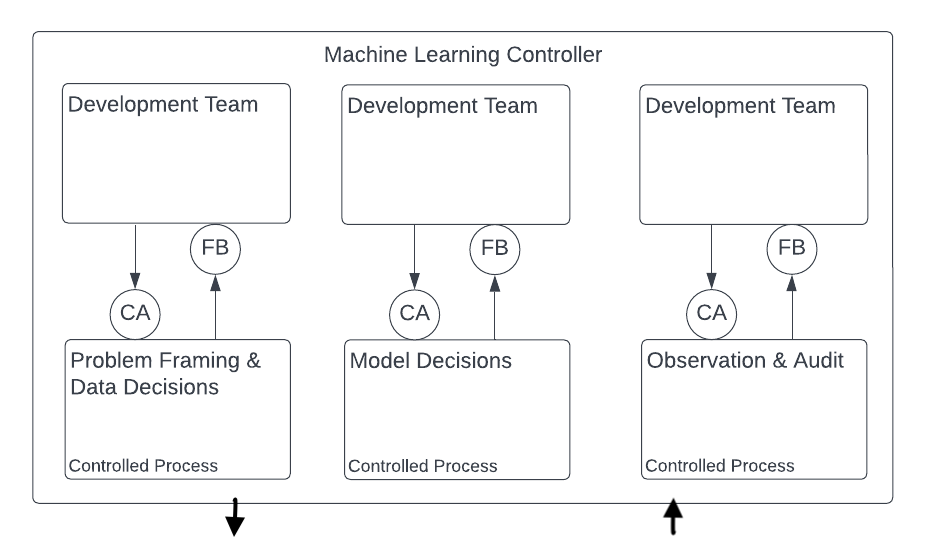}
         \caption{Modeling Life-Cycle Control}
         \label{dev}
         \label{cycle-treatment}
\end{figure}
  

\begin{figure}[h!]
    \centering
    \includegraphics[width=0.49\linewidth, height=6cm]{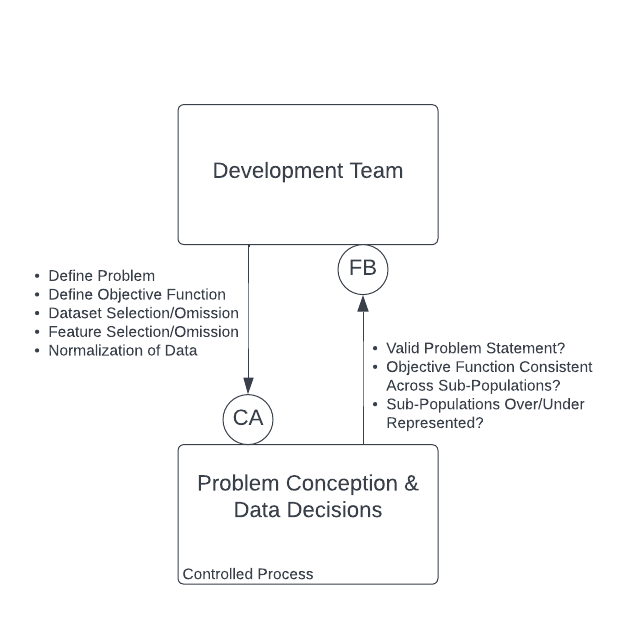}
    \caption{Data Decisions}
    \label{data}
\end{figure}

\subsection{STAMP and STPA}
Accepting a complex systems viewpoint still leaves you with the problem of how to take kind of linear engineering practices and capture. All the complexity of nonlinearity of problem. And the big insight of stamp is to say the complexity arises from the structure of the system and you can consider the structure in a linearized way and that gets you alot further.
\section{Losses and Hazards}
In our initial analysis we identified over thirty potential social, ethical and safety losses that could result both from the PDMP score system proper or downstream and yet a result of the PDMP score system's interactions within the broader healthcare system, these are fully outlined in Appendix \ref{initial_loss}. To simplify analysis going forward it was then necessary to reduce this loss list to five general categories as defined in Appendix \ref{distilled}.  
\begin{enumerate}
    \item Patient Death
    \item Inequity between social groups
    \begin{enumerate}
        \item Allocative - Disparity in PDMP risk score can result in a disparity in: 
        \begin{itemize}
            \item Health Treatment: affecting subsequent opportunity as well if resulting treatment disparity is debilitating reducing ability to hold a job or care for children or adult dependents.
           \item Job Opportunity: Is PDMP risk score specifically prohibited from being considered when seeking a drug-dispensing or other related health care job? Can the score or some subset be an input to other products such as background checking systems, credit or hiring algorithms?
        \end{itemize}
        \item Representational - A grouping with an inappropriately high score may have the effect of categorizing a patient inappropriately as more likely drug-seeking.
        \item Quality of Service - More difficult interactions, extra intrusive questions, when interacting with Doctors, health staff and pharmacies. 
        \begin{enumerate}
            \item Alienation:
            \begin{itemize}
                \item Turned away at Pharmacies: Resulting in adverse emotions, distrust and exclusion from the benefits offered others for health treatment.
                \item Turned away as a new patient: same as above
                \item Dismissed as a patient: same as above
            \end{itemize}
            \item Increased Labor: Above reasons in Alienation repeated here as all result in additional labor for the patient to overcome to get appropriate treatment.
	        \item Service or Benefit Loss: for same reasons in alienation, benefit of treatment is lost when it cannot be overcome or cost/effort required is too high to fight.
        \end{enumerate}
    \end{enumerate}
    \item Patient has untreated pain - Physical, mental anguish, social damage, 
    \begin{enumerate}
        \item Mental Health
        \item Physical debilitation
        \item Social Damage
        \item Occupation Damage
        \item Family Care Damage
    \end{enumerate}
    \item Loss of Safe access to Treatment/Care (Abandonment): 
    \item Behavior Herding: Desperate, deeply affected individuals may be herded to get the care they need from illegal means, thereby increasing risk of incarceration, addiction, abuse and death as the illegal treatment has no protections from overdose or doctor and pharmacist oversight.
    \item Loss of patient care (narcotic, benzodiazepines, stimulants; overall)
    \item Degraded Quality of Life: Loss of ability to work, care for children, enjoy normal life, care for adult dependents. 
    \item Law enforcement action - See CA state review -- ``law enforcement surveillance and its attendant threat of criminal investigation and prosecution incentivize patient abandonment, forced taper, and involuntary medication discontinuation.
    \item Reputation loss 
    \item Privacy Violations
    \item Licensure (Doc/Pharm)
    \item Increased Liability Insurance (Docs and Pharm)
    \begin{enumerate}
        \item Social Control
        \item Financial
    \end{enumerate}
    \item Loss of Autonomy, clinical judgment 
    \item Inequity with social groups (poor people may have higher scores given method of payment is a factor)
    \begin{enumerate}
        \item Sexual assault survivor
        \item Prior arrest history
        \item Age, socioeconomic, regional, race, gender, sexuality
    \end{enumerate}
    \item Reduced accessibility to Doctors
\end{enumerate}

\subsection{Reduced Loss List}\label{distilled}
\begin{enumerate}
    \item Death, Injury or Disability:
    \begin{itemize}
        \item Patient Death
        \item Untreated Medical Conditions (Pain)
        \item Additional Physical or Mental Injury
    \end{itemize}
    \item Disparity of Benefit/Harm
    \begin{itemize}
        \item Allocative Disparity
        \item Representational Disparity
        \item Quality of Service Disparity
    \end{itemize}
    \item Social/Economic Injury
    \begin{itemize}
        \item Damage to Reputation
        \item Occupational Damage
        \item Family Damage 
        \item Privacy Violations
    \end{itemize}
    \item Damage to Quality of Healthcare
    \begin{itemize}
        \item Abandonment
        \item Loss of Autonomy in Clinical Judgement
        \item Loss of Opportunity for Care, i.e., reduced accessibility to Doctors
    \end{itemize}
    \item Coerced Criminality or Unsafe Treatment
    \begin{itemize}
        \item Herding to Unsafe/Illicit Behavior
        \item Increase in Law Enforcement Scrutiny
    \end{itemize}
\end{enumerate}

\subsection{PDMP Score Hazards}\label{hazards}
This table lists all of the identified PDMP Scoring system hazards cross-referenced with the potential losses which may result from those hazards.
\begin{table}[h!]
\begin{tabular}{ |p{0.25cm}|p{6cm}||p{0.5cm}|p{0.5cm}|p{0.5cm}|p{0.5cm}|p{0.5cm}|  }
 \hline
 \multicolumn{7}{|c|}{PDMP Risk Score Hazards} \\
 \hline
 &Hazards&L1&L2&L3&L4&L5\\
 \hline
 1&Over-prescribe   &X&&&X&\\
 \hline
 2&Under-prescribe &X&&X&X&\\
 \hline
 3A&Inappropriately Scored - High &X & &X &X&X\\
 \hline
 3B&Inappropriately Scored - Low &X & &  &X&\\
 \hline
 4&Score Leaked &X&X&X&X&X\\
 \hline
 5&Problematic/Biased Data & &X&&X&\\
 \hline
 6&Abandonment &X&&X&X&X\\
 \hline
 7&Not provided most effective treatment.&X&&X&X&X \\
 \hline
 8&Patient gives up on medical system.&X&&X&&X \\
 \hline
 9&Excessive false positives.&X&X&X&X& \\
 \hline
 10&Excessive false negatives.&X&&&X& \\
 \hline
\end{tabular}
\caption{PDMP Risk Score System Hazards}\label{haz_tbl_PDMP}
\end{table}

\section{PDMP Score Unsafe Control Actions}\label{UCA}
\begin{table}[h!]
\begin{tabular}{ |p{1.25cm}||p{3cm}|p{2.5cm}|p{0.5cm}|p{3cm}|p{3cm}|  }
 \hline
 \multicolumn{6}{|c|}{UCAs: Patient Care Control Loop} \\
 \hline
 Control Action&Not Provided&Provided&TE TL&Too Low&Too High\\
 \hline
Risk Score & Score defaults to zero. Hazard if patient susceptible to addiction - H1&H6, H7, H8&N/A&H1, H3, H10&H2, H3A, H6, H7, H8, H9\\

 \hline
\end{tabular}
\caption{Patient Care: Unsafe Control Actions}
\label{UCA_care}
\end{table}

\begin{table}[]
\begin{tabular}{ |p{3cm}||p{0.5cm}|p{2.5cm}|p{0.5cm}|p{3cm}|p{3cm}|  }
 \hline
 \multicolumn{6}{|c|}{UCAs: Problem Conception and Data Decision Control Loop} \\
 \hline
 Control Action&Not Prov&Provided&TE TL&Too Few/Little&Too Many/Much\\
 \hline
Define Problem &N/A&H5, H3&N/A&N/A&N/A\\
Define Obj. Function&N/A&H1-3, H5-10&N/A&N/A&N/A\\
Dataset Selection or Omission&N/A&H1-3, H5-10&N/A&N/A&N/A\\
Feature Selection or Omission&N/A&H1-3, H5-10 &N/A&H1-3,H6-10 &H1-3,H6-10\\
Data Normalization&N/A&H3, H5, H9, H10&N/A&H3, H5, H9, H10&H3, H5, H9, H10\\
 \hline
\end{tabular}
\caption{Problem Conception and Data Decisions}
\label{UCA_dev}
\end{table}

\subsection{Facial Recognition Technology Losses}\label{frt_losses}
This table lists all of the Facial Recognition system losses, both societal and individual.

\begin{table}[h!]
\centering
\begingroup
\setlength{\tabcolsep}{10pt} 
\renewcommand{\arraystretch}{1.5} 
\begin{tabular}{ |p{0.25cm}|p{10cm}| }
 \hline
 \multicolumn{2}{|c|}{Facial Recognition System Losses} \\
 \hline
 &\textbf{Societal} \\
 \hline
 1&Loss of Justice 
   \setlist{nolistsep}
   \begin{itemize}[noitemsep]
   \item Wrongful Convictions or No Convictions
   \item Loss of Confidence in Overall System \newline by Population or Demographic
   \item Loss of Public Service Reputation
   \end{itemize}
   \\
 \hline
 2&Loss of Public Service Resources \newline (i.e. Manpower, Money, Time, Equipment) \\
 \hline
 3&Loss of Opportunity to Population or Demographic \\
 \hline
 4&Loss of Privacy: Surveillance State or Perception Thereof \\
 \hline
 \hline
 &\textbf{Individual} \\
 \hline
 1& Loss of Life / Physical Harm / Harassment \\
 \hline
 2& Loss of Opportunity \newline (e.g. Missing out on College, Scholarship, or Job due to Wrongful Arrest) \\
 \hline
 3& Loss of Resources (Time/Money/Effort) \\
 \hline
 4& Loss of Confidence in Justice System \\
 \hline
 5& Loss of Privacy / Autonomy - Surveillance \\
 \hline
\end{tabular}
\endgroup
\caption{Facial Recognition System Losses}\label{haz_tbl}
\end{table}

\subsection{Facial Recognition Technology Hazards}\label{frt_hazards}
This table lists all of the identified Facial Recognition system for law enforcement hazards cross-referenced with the potential losses which may result from those hazards.
\begin{table}[h!]
\centering
\begingroup
\setlength{\tabcolsep}{10pt} 
\renewcommand{\arraystretch}{1.5} 
\begin{tabular}{ |p{0.25cm}|p{6cm}|p{0.5cm}|p{0.5cm}|p{0.5cm}|p{0.5cm}|p{0.5cm}|p{0.5cm}|p{0.5cm}|p{0.5cm}|p{0.4cm}| }
 \hline
 \multicolumn{11}{|c|}{Facial Recognition System Hazards} \\
 \hline
 &Hazards&S1&S2&S3&S4&I1&I2&I3&I4&I5 \\
 \hline
 1&Wrongful Warrant Issuance   &X&X&&&X&X&X&X&X \\
 \hline
 2&Investigator Cognitive Bias (i.e. automation bias, anchoring)   &X&X&X&X&X&X&X&X&X \\
 \hline
 3&Leaked Data (Identities)  &&X&&&X&X&X&X&X \\
 \hline
 5&No Guidance on FR use. Lax Control of FR confidence level - Low &X&X&&X&X&X&X&X&X \\
 \hline
 6&Investigator Improper Confidence in Results &X&X&X&&X&X&X&X&X \\
 \hline
 7&Face Detection Evasion &X&&&&X&X&X&X&X \\
 \hline
 8&Unverified Probe Photo Alteration \newline (i.e. Photoshop) &X&X&&&X&X&X&X&X \\
 \hline
 9&Operator Error / Training Deficiency &X&X&X&X&X&X&X&X&X \\
 \hline
 10&Unauthorized/Unintended/Unethical use of FRT (i.e. surveillance) &X&X&X&X&X&X&X&X&X \\
 \hline
 11&Actual Perpetrator Not in Database &X&X&X&&X&&&X& \\
 \hline
 12&False Match (False Positive) &X&X&&&X&X&X&X&X \\
 \hline
 13&False Non-Match (False Negative) &X&X&&&X&&&X& \\
 \hline
 14&Different Performance of FR Between Demographics &X&&X&X&&&&& \\
 \hline
 15&DB Tampering (Ingested Photos) &X&X&X&X&X&X&X&X&X \\
 \hline
 16&Frame an Individual through Matching Evasion Attack &&&&&X&X&X&X&X \\
 \hline
 17&Sparse Photo Representation for Identity &X&X&X&&X&&&&X \\
 \hline
 18&Mislabeled Photos in DB &X&X&&&X&X&X&X& \\
 \hline
 19&Identity Mismatch in DB &&&&&X&X&X&X&X \\
 \hline
 20&Face Not Detected or Wrong Portion of Image Captured as Face &X&&&X&&&&&X \\
 \hline
 21&Reinforcing Classification Feedback Loops &&&X&&&&&& \\
 \hline
 22&Stale DB Images or Embeddings of a Subject(s) &X&X&&&X&X&X&X&X \\
 \hline
\end{tabular}
\endgroup
\caption{Facial Recognition System Hazards}
\end{table}

\section{FRT UCAs}
\begin{table}[!ht]
    \centering
    \begingroup
    \setlength{\tabcolsep}{5pt} 
    \renewcommand{\arraystretch}{1.5} 
    \begin{tabular}{|p{2.5cm}|p{3.2cm}|p{3.2cm}|p{3.2cm}|p{3.2cm}|}
    \hline
 \multicolumn{5}{|c|}{UCA 1 - Problem Framing and Data Decisions} \\
        \hline
        Control Action & Not Providing & Providing & Too Early/Too Late\newline (Wrong Timing) & Too Low (TL) /\newline Too High (TH) \\ 
        \hline
        (1) Define Problem & Not beginning by clearly defining the problem to solve is hazardous. \newline {\footnotesize[H6,10,14,17,20]} &  & Defining the problem after a model is built (too late) is hazardous.\newline {\footnotesize[H6,10,14]} & \\ \hline
        (2) Define Social and Ethical Success Metrics (What is Good) & Not providing success metrics that consider social and ethical impacts is hazardous in that it can result in a mismatch between model performance and desire, i.e., Robustness to evasion attack. impacts \newline {\footnotesize[H1,6,7,10,12,13,14, 20]} & Providing insufficient success metrics will result in hazardous states. \newline {\footnotesize [same]} & & \\ \hline
        (3) Dataset Acquisition (Selection or Omission) & Not providing dataset selectivity criterion may result in a hazardous states as it means ingestion is wide open. \newline {\footnotesize[H6,12,13,14,15,16,18,\newline 19,21]} & Providing Dataset Selectivity may result in hazardous states by eliminating needed data. \newline {\footnotesize [H11,17]} & Being selective too late is hazardous as it may result in unethical or illegal data inclusions in training the model. \newline {\footnotesize[H18,H19,H20,H21]} & Being either too restrictive or insufficiently restrictive is hazardous.  \newline {\footnotesize[TL: H14,15,16,18,21]} \newline {\footnotesize[TH: H6,11,12,13,17]} \\ \hline
        (4) DB Update Policy (When and at what periodicity) & Not providing for CL is hazardous in that the model becomes stale, new adults are not added, aging faces not accounted for etc. \newline {\footnotesize[H6,11,12,13,17,22]} & Providing for CL is hazardous in that you introduce a number of potential harms and avenues for attack. \newline {\footnotesize[H7,11,12,13,14,15,16,21]} &  & Individual Update Rate \newline {\footnotesize[TL: H6,12,13,22]} \\ \hline
        (5) Source Weighting & Not providing is hazardous as it may make it easier to supplant highly trusted data sources with less trusted data. \newline {\footnotesize[H15,16]} &  &  &   \\ \hline
        (6) Data Processing (e.g., face normalization, data balancing: by individual or population, etc) & Not Providing appropriate data pre-ingestion data processing and balancing is hazardous. \newline {\footnotesize[H11,12,13,14,17]} & These techniques are not perfect and may introduce additional error where and may do so in disproportionate ways if they are more or less effective based on race, gender or other demographic. \newline {\footnotesize[H12,13,14]} & &\\ \hline
    \end{tabular}
    \endgroup
    \caption{Facial Recognition Problem Framing and Data Decisions Unsafe Control Actions}\label{CJFR_UCA1}l
\end{table}

\begin{table}[!ht]
    \centering
    \begingroup
    \setlength{\tabcolsep}{5pt} 
    \renewcommand{\arraystretch}{1.5} 
    \begin{tabular}{|p{2.5cm}|p{3.2cm}|p{3.2cm}|p{3.2cm}|p{3.2cm}|}
    \hline
 \multicolumn{5}{|c|}{UCA 2 - Model Decisions} \\
        \hline
        Control Action & Not Providing & Providing & Too Early/Too Late\newline (Wrong Timing) & Too Low (TL) /\newline Too High (TH) \\ 
        \hline
        (1) Architecture: Complexity(Model) &  &  &  & Both DNN complexity too low and too high are hazardous. Seek through sufficiently representative Validation Sets to verify not over or under fitted. {\footnotesize[H12,13,14]}\\ \hline
        (2) Properly Define/Modify Objective Function/Loss and Optimization Procedure & Given FB (Performance Measures) not modifying {\footnotesize[H12,13,14]} \newline Failure to properly match and modify(eg, soft F-beta or class reweighting) objective function is hazardous. {\footnotesize[H6,12,13,14,17]} & & & \\ \hline
        (3) Allowed / Recommended Threshold and K determinations & & & & If k or threshold is too low, will result in hazardous states high non-match {\footnotesize[H13]} , too high may result in high false match rate {\footnotesize[H12]} \\ \hline
        (4) Data Augmentation Strategy & Not providing can be hazardous. \newline {\footnotesize[H12,13,14,17]} & & & \\ \hline
        (5) Defense Techniques to Defend against Adversarial attacks (evasion) & Not providing is hazardous. {\footnotesize[H7]} & {\footnotesize[H12,13]} &  &   \\ \hline
    \end{tabular}
    \endgroup
    \caption{Facial Recognition Model Decisions Unsafe Control Actions}\label{CJFR_UCA2}l
\end{table}

\begin{table}[!ht]
    \centering
    \begingroup
    \setlength{\tabcolsep}{5pt} 
    \renewcommand{\arraystretch}{1.5} 
    \begin{tabular}{|p{2.5cm}|p{3.2cm}|p{3.2cm}|p{3.2cm}|p{3.2cm}|}
    \hline
 \multicolumn{5}{|c|}{UCA 3 - Deployment, Observation and Audit Phase} \\
        \hline
        Control Action & Not Providing & Providing & Too Early/Too Late\newline (Wrong Timing) & Too Low (TL) /\newline Too High (TH) \\ 
        \hline
        (1) Establishing a Risk Tolerance for Performance and Social and Ethical Impact Considerations 
        & Not providing is hazardous as it does not enable thresholds to trigger meaningful actions to prevent losses. \newline {\footnotesize[H1,6,7,10,12-22]} &  & & Risk Tolerance that is either too high or too low is hazardous, if too low will lead to {\footnotesize[H12,13]} too high can result in all those under not providing.\newline {\footnotesize[H6,7,10,12-22]} \\ \hline
        (2) Load Testing, Internal Verification and Validation of critical go/No Go Criteria. 
        & Not determining and validating critical go/no go criteria is hazardous \newline {\footnotesize[H1,6,10,12-14]} & & & \\ \hline
        (3) Perform DB and Model Update based on requirements defined in Data Decisions. & Not providing is hazardous as it will mean your DB of images and associated embeddings to compare probe photos to will become stale as humans age or reach age of majority. If it is not performed in a balanced way it can also be hazardous \newline {\footnotesize[H1,6,11-14,17,22]} & Providing can be hazardous \newline {\footnotesize [H15,16,18,19,21]} & & Too low, (not frequently enough), is hazardous for same reason as not providing. {\footnotesize[H1,6,11-14,17,22]} Too high(too frequently) is hazardous as it increases the opportunity for those in providing. {\footnotesize[H15,16,18,19,21]} \\ \hline
        (5) Continuous and/or Periodic Monitoring of Social and Ethical Success Metrics (Reference Problem Framing CA2) 
        & Not providing is hazardous because, if the assumptions of the model and system do not hold, a resulting change in Social and Ethical performance will not be detected and corrected. \newline {\footnotesize[H1,6,10,12-14]} & & & Too low, (not frequently enough), is hazardous for same reasons as not providing. \newline {\footnotesize[H1,6,10,12-14]} \\ \hline
        (6) Thresholds for alert and/or Corrective Actions 
        & Not providing is hazardous as the system will not take action when the risk limits are surpassed. \newline {\footnotesize[H1,6,12-14,20]} &  &  & Too low risks alarm fatigue, too high risks those in not providing.   \\ \hline
    \end{tabular}
    \endgroup
    \caption{Deployment, Observation and Audit Unsafe Control Actions}\label{CJFR_UCA3}l
\end{table}

\section{ML Harms} \label{harms}
A steadily growing number of incidents and calls to action demonstrate the necessity to include social and ethical impact analysis as a key component of the ML system development life cycle~\cite{bender_dangers_2021,politico_dutch_2022,perkowitz_bias_2021,Angwin_arrest_recidivism_2016}. However, in order to enable such an assessment of social and ethical impacts, we must first begin with a firm understanding of the various harms that can result from sociotechnical algorithmic systems. Additionally, special attention must be given to the deployment environment as sociotechnical systems often have far reaching social and technological connections and impacts for humans which can result in losses, hazards, and negative outcomes for entities far removed from the system--in the case of PDMP scoring systems these will no doubt include patients, doctors, and pharmacists, but also patient families and the functionality of and trust in the health system at large. 

This paper adopts Shelby \textit{et al}.'s recent work which successfully taxonomized the myriad manifestations of algorithmic harms~\cite{shelby2023harms}. This taxonomy provides an initial yet robust foundation to proceed from, and we use it as a basis for developing our subject system's social and ethical losses and hazards, a key part of the first step of STPA.

\section{Conducting STPA for ML S\&E Impact: PDMP}\label{execute_pdmp_stpa}
Our STPA results initially identified 30 losses which we pared down to five loss types (Appendix \ref{distilled}). Progressing further, we identified 11 hazardous system states and in table \ref{haz_tbl_PDMP} mapped each of these to related losses. Next, the team outlined the sociotechnical context and control structure (see figure \ref{PDMP_ctrl}) where the analysis effort was scoped to the two most impactful of the seven primary control loops, namely \textit{Patient Care}, figure \ref{care} and \textit{Data Decisions}, figure \ref{data}. Analysis resulted in 13 unsafe control actions (UCAs) as shown in tables \ref{UCA_care} and \ref{UCA_dev}, respectively. Once the UCAs are captured in the analysis table and linked to hazards, the developers have a ready made list of actionable sociotechnical system requirements for fair and ethical treatment complete with traceability back to stakeholder defined unacceptable losses. Lastly, examining loss scenarios and thus each UCA helps system owners, operators, and developers systematically eliminate or mitigate hazardous systems states tied to identified harms. 
You can have as much text here as you want. The main body must be at most $8$ pages long.
For the final version, one more page can be added.
If you want, you can use an appendix like this one, even using the one-column format.

\end{document}